%% file: 2010swat4ls.tex
\documentclass{llncs}
\usepackage{makeidx}  
\usepackage{paralist} 
\usepackage{lineno} 
\usepackage{graphicx} 
\usepackage[colorlinks=false]{hyperref}
\usepackage{wrapfig} 
\usepackage{floatflt}
\usepackage[font=small,format=plain,labelfont=bf,up,textfont=it,up]{caption}
\usepackage{setspace}
\usepackage{amssymb}
\usepackage{MnSymbol}
\usepackage{stmaryrd}
\usepackage{paralist}
\usepackage{listings}

\DeclareMathSizes{8}{8}{8}{8}

\newcommand{\ie}{i.e.~\/}

\newcommand{\etal}{\textit{et al~\/}}
\newcommand{\etalnospace}{\textit{et al}}
\newcommand{\cabignospace}{caBIG\textsuperscript{\textregistered}}
\newcommand{\cabig}{caBIG\textsuperscript{\textregistered}~\/}
\newcommand{\cabiglong}{cancer Biomedical Informatics Grid\textsuperscript{\textregistered}~\/}
\newcommand{\comment}[1]{}

\newcommand{\arrow}{$\rightarrow$}
\newcommand{\semicolon}{$|$}
\newcommand{\comma}{}
\newcommand{\terminal}[1]{{\scriptsize \begin{bf}#1\end{bf}}}
\newcommand{\nonterminal}[1]{$\langle${\scriptsize #1}$\rangle$}
\newcommand{\emptystring}{$\epsilon$~\/}


\begin{document}

\lstdefinestyle{numbers}
{numbers=left, stepnumber=1, numberstyle=\tiny, numbersep=10pt}
\lstdefinestyle{nonumbers}
{numbers=none}

\frontmatter          
\pagestyle{headings}  
%

\mainmatter              
%
\title{Ontology-based Queries over Cancer Data}
\titlerunning{Ontology-Based Queries over Cancer Data}  
%
\author{Alejandra Gonz\'{a}lez-Beltr\'{a}n\inst{1}$^{,}$\inst{2} \and Ben Tagger\inst{1}
\and Anthony Finkelstein\inst{1}}
\authorrunning{AGB et al.} 
%
%
\institute{Department of Computer Science 
\and
Computational and Systems Medicine\\ University College London, London WC1E 6BT, United Kingdom }

\maketitle              

\begin{abstract}
\input{abstract}
\end{abstract}

\section{Introduction}
\label{section:introduction}
\input{introduction}

\section{Background}
\label{section:background}
\input{background}

\section{Analysis of the caGrid Query Language}
\label{section:cqlanalysis}
\input{cqlanalysis}

\section{Ontology-based queries over the caGrid infrastructure}
\label{section:ontoqueries}
\input{ontoqueries}

\subsection{OWL Representation of caGrid Information Models}
\label{section:owlrep}
\input{owlrep}

\subsection{Query Rewriting and Translation}
\label{section:querytrans}
\input{querytrans}

\subsection{Implementation}
\label{section:implementation}
\input{implementation}

\subsection{Performance Evaluation}
\label{section:perfeval}
\input{perfeval}

\section{Related Work}
\label{section:relatedwork}
\input{relatedwork}

\section{Conclusions}
\label{section:conclusions}
\input{conclusions}

\subsubsection{Acknowledgements}
\input{acknowledgements}

%
%
\begin{spacing}{0.9}
\bibliographystyle{unsrt}
\bibliography{biblio}
\end{spacing}

\comment{
\clearpage
\addtocmark[2]{Author Index} 
\renewcommand{\indexname}{Author Index}
\printindex
\clearpage
\addtocmark[2]{Subject Index} 
\markboth{Subject Index}{Subject Index}
\renewcommand{\indexname}{Subject Index}
\input{subjidx.ind}
}
\end{document}

%% file: abstract.tex
The ever-increasing amount of data in biomedical research, and in cancer research in particular, needs to be managed to support efficient data access, exchange and integration. Existing software infrastructures, such \emph{caGrid}, support access to distributed information annotated with a domain ontology. However, caGrid's current querying functionality depends on the structure of individual data resources without exploiting the semantic annotations. In this paper, we present the design and development of an ontology-based querying functionality that consists of: the generation of OWL2 ontologies from the underlying data resourcesÕ metadata and a query rewriting and translation process based on reasoning, which converts a query at the domain ontology level into queries at the software infrastructure level. We present a detailed analysis of our approach as well as an extensive performance evaluation. While the implementation and evaluation was performed for the \emph{caGrid} infrastructure, the approach could be applicable to other model and metadata-driven environments for data sharing.

%% file: introduction.tex
In the biomedical sciences, the use, exchange and integration of the ever-increasing amount of data has become paramount to accelerate the discovery of new approaches for the detection, diagnosis, treatment and prevention of diseases. In particular, this applies to cancer, for which the US National Cancer Institute (NCI) and the UK National Cancer Research Institute (NCRI) have implemented the \cabignospace\footnote{\cabig stands for \cabiglong} programme and the NCRI Informatics Initiative, looking at building and deploying software infrastructure to manage and analyse data generated from heterogenous data sources.

In this paper, we provide an analysis of the caGrid\cite{saltz:Bioinformatics2006} software infrastructure developed within the NCI \cabig programme and extend it with richer querying capabilities. caGrid supports a collaborative information network for sharing cancer research data, and deals with syntactic and semantic interoperability of the data resources in a service-oriented model-driven architecture. Semantic interoperability is achieved by using a metadata registry, which maintains information models annotated with concepts from a domain ontology: the NCI thesaurus (NCIt)\cite{hartel:JBI2005}. However, the query functionality provided in caGrid does not take into account the semantic annotations, but it only relies on each individual information model. 

Our methodology is based on extending the caGrid service-oriented model-driven infrastructure with additional services to support ontology-based queries over the distributed data resources. In this way, the biomedical researchers, as the end-users of our system, will be able to query cancer data by building queries using their domain knowledge (expressed as concepts from the NCIt ontology) rather than having to know the underlying models. This also means that the queries are reusable across resources, which is not the case in the caGrid infrastructure. This functionality will be incorporated into the NCRI ONcology Information eXchange (ONIX\footnote{\href{http://www.ncri-onix.org.uk/}{http://www.ncri-onix.org.uk/}}). Our approach involves a customised transformation from annotated information models to an ontological representation using the Web Ontology Language version 2 (OWL\footnote{OWL is a recommendation from the World Wide Web Consortium (W3C) and the language overview for its second version can be found at \href{http://www.w3.org/TR/owl2-overview/}{http://www.w3.org/TR/owl2-overview/}}). This representation supports annotations based on a primary concept and a list of qualifiers. Based on these ontological representations of the data resources, we have designed and developed a query rewriting and translation approach that converts concept-based queries into the query language supported by the caGrid infrastructure. This approach is general and could be used to support other target query languages, as the only step dependent on caGrid is the last one. This work presents significant improvements over our previous work\cite{gonzalez-beltran:CBMS2009}, as we have significantly modified and improved the OWL representation and the design and implementation of the query rewriting and translation steps. We have developed a caGrid analytical service for the transformation from an annotated information model to OWL. Additionally, we present an analysis of the caGrid query language and information together with an extensive performance evaluation that justifies the applicability of our solution. 

This paper is structured as follows. Section \ref{section:background} introduces background material on the caGrid infrastructure. Section \ref{section:cqlanalysis} presents an analysis of the caGrid query functionality and the type of queries supported by its query language. Then, we present in section \ref{section:owlrep} the OWL representation that is used for query rewriting and translation, which in turn is described in Section \ref{section:querytrans}. The implementation details and performance evaluation results are given in Sections \ref{section:implementation} and \ref{section:perfeval}, respectively. The evaluation includes an analysis of the generated ontologies as well as several performance metrics for OWL generation and query rewriting, which justify the viability of our approach. After comparing our approach with related work in Section \ref{section:relatedwork}, we conclude the paper in Section \ref{section:conclusions}, including considerations for future work.

%% file: background.tex
\newcounter{fnnumber}
\cabignospace\cite{saltz:Bioinformatics2006} is an NCI programme whose aim is to create a virtual and federated informatics infrastructure for sharing data, tools and connect scientists and organisations in the cancer research community. The computing middleware in \cabig is called caGrid, which is a Grid\cite{foster:anatomy01} extended to support data modelling and semantics\cite{saltz:Bioinformatics2006}. caGrid has a number of core services and corresponding application programming interfaces (APIs), which we will introduce next, by analogy with the metadata hierarchy\cite{pollock:adaptive04}, as per Figure \ref{fig:cagrid}.

\begin{floatingfigure}[r]{80mm}
	\begin{center}
  \includegraphics[width=78mm]{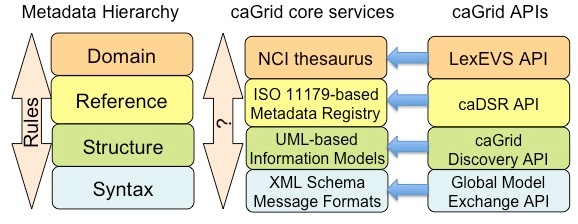}
	\end{center}
	\caption{caGrid semantic infrastructure. \label{fig:cagrid}}
\end{floatingfigure}

The metadata hierarchy represents how the semantics of raw data (\textit{instance data}) can be augmented by overlaying metadata of increasing descriptiveness \cite{pollock:adaptive04}. The \textit{syntactic metadata} refers to the language format and data types, and in caGrid is represented by XML schemas managed by the Global Model Exchange (GME)\cite{saltz:Bioinformatics2006} service, which exposes them through the GME API\footnote{\href{http://cagrid.org/display/gme/}{http://cagrid.org/display/gme/}}. The \textit{structural metadata} gives form to the units of data. In caGrid, it is implemented as an object-oriented virtualisation of the underlying data resources\cite{saltz:Bioinformatics2006} and it is represented as UML\footnote{UML stands for the Unified Modeling Language, a specification of the Object Management Group\textregistered (OMG\textregistered)} models. These UML models can be accessed through the Discovery API\footnote{\href{http://cagrid.org/display/metadata13/Discovery}{http://cagrid.org/display/metadata13/Discovery}}. The purpose of the \textit{referent metadata} is to represent the linkages between the different data models. In caGrid, the linkages are provided by a metadata registry, called caDSR\footnote{caDSR stands for cancer Data Standards Repository}, based on the ISO/IEC 11179 standard\footnote{\href{http://metadata-stds.org/11179/}{http://metadata-stds.org/11179/}}. caDSR manages common data elements (CDEs) and exposes them through the caDSR API. The \textit{domain metadata} represents what the data is about. It is implemented by a domain conceptualisation, usually in the form of an ontology\cite{pollock:adaptive04}. In the caGrid case, the NCIt ontology\cite{hartel:JBI2005} is used, accessed via the LexEVS API\footnote{\href{https://cabig.nci.nih.gov/concepts/EVS/}{https://cabig.nci.nih.gov/concepts/EVS/}}. Finally, the rules constitute an overarching layer that can be applied to all the aforementioned layers. Rules can be used to constrain and extend the semantics of metadata specifications at any of the abstraction levels\cite{pollock:adaptive04}. In the current caGrid semantic infrastructure, there is no equivalent to the \textit{rule metadata}.

A data resource owner can share the data by developing caGrid data services using common interfaces and metadata, as described above. In this way, a data service encapsulates the data, which is kept in native formats (including, for example, relational data or flat files), exposing an access interface based on the object-oriented (UML) model of the underlying resource. The common interface also exposes a query processor based on the Common Query Language (CQL) defined for caGrid. CQL is an object-oriented query language reflecting the underlying object model of the data resource while abstracting the physical representation of the data\cite{saltz:Bioinformatics2006}. At the time of writing, there exist two versions of CQL and there is a pre-release version of the latest one\footnote{\href{http://cagrid.org/display/dataservices/CQL+2}{http://cagrid.org/display/dataservices/CQL+2}}. More details on CQL are given in Section \ref{section:cqlanalysis}.

caGrid also supports basic distributed aggregations and joins of queries over multiple data services by means of the caGrid Federated Query Infrastructure\footnote{\href{http://cagrid.org/display/fqp/Home}{http://cagrid.org/display/fqp/Home}}, through a distributed extension of CQL called DCQL. Thus, caGrid relies on D/CQL -- custom query languages based on the structural characteristics of the resources. In other words, caGrid builds a 'structural layer', where queries are expressed in terms of objects, attributes and associated objects, without allowing for semantic queries.  D/CQL are evolving to provide richer structural queries as new requirements arise from different \cabig projects. However, these query languages do not allow for data extraction based on semantic information. Thus, a shortcoming of caGrid is that does not currently exploit the \textit{referent} and \textit{domain metadata} maintained for its data services. Additionally, as already mentioned, it is not possible to specify \textit{rules} about the models nor the domain semantics.

As stated in the introduction, this work advocates the extension of the caGrid infrastructure to exploit its rich metadata by building a semantic layer, using semantic web technologies to exploit caGrid's metadata. Additionally, this extension is capable of: \begin{inparaenum}[\itshape a\upshape)] \item accommodating other resources with different ways of dealing with metadata, and \item specifying rules at different levels of abstraction.\end{inparaenum}

%% file: cqlanalysis.tex
A CQL query is defined by an XML document, which must comply to a specified XML schema\footnote{The CQL XML schema is available at: \href{http://cagrid.org/display/dataservices/CQL+Schemas}{http://cagrid.org/display/dataservices/CQL+Schemas}}. The schema indicates that a CQL query must specify a {$\langle$}Target{$\rangle$} element, which is the data type of the query result. Optionally, an {$\langle$}Attribute{$\rangle$} element might indicate a predicate over an attribute of the object with {$\langle$}Target{$\rangle$} type and an {$\langle$}Association{$\rangle$} may specify a link with a related object. In Table \ref{table:cqlgrammar} we show how a CQL query is built recursively presenting it as a context-free grammar, where {$\langle$}CQLQuery{$\rangle$} is the start symbol, \emptystring is the empty string and \nonterminal{xsd:string} is the non-terminal variable representing the {\emph XSD:string} data type.

\begin{table}[ht]
\begin{tabular}{c}
\begin{minipage}[b]{0.499\linewidth}\centering
\begin{tabbing}
 \nonterminal{CQLQuery} \arrow \= \nonterminal{Target} \semicolon \\
                       \>  \nonterminal{Target} \comma \nonterminal{QueryModifier}\\
\nonterminal{Target} \arrow \= \nonterminal{Name} \comma \nonterminal{Attribute} \semicolon \\
           					      \> \nonterminal{Name} \comma \nonterminal{Association} \semicolon \\
					             \> \nonterminal{Name} \comma \nonterminal{Group} \\					
\nonterminal{Attribute} \arrow \nonterminal{Name} \comma \nonterminal{Predicate} \comma \nonterminal{Value}\\
\nonterminal{Group} \arrow  \= \nonterminal{LogicalOp} \comma \nonterminal{Attribute} \comma \nonterminal{Group1} \semicolon \\
		\> \nonterminal{LogicalOp} \comma \nonterminal{Association} \comma \nonterminal{Group1}  \\
\nonterminal{Group1} \arrow  \= \nonterminal{Attribute} \comma \nonterminal{Group1}  \semicolon \\
				                    \> \nonterminal{Association} \comma \nonterminal{Group1}  \semicolon \\
		                                  \> \nonterminal{Group} \semicolon \emptystring \\
\nonterminal{Name} \arrow \nonterminal{xsd:string} \\
\nonterminal{RoleName} \arrow \nonterminal{xsd:string} \\

\end{tabbing}
\end{minipage}
\end{tabular}
\hspace{0.5cm}
\begin{tabular}{c}
\begin{minipage}[b]{0.499\linewidth}
\begin{tabbing}			  
\nonterminal{LogicalOp} \arrow \terminal{AND} \semicolon \terminal{OR}\\
\nonterminal{Predicate} \arrow \= \terminal{EQUAL\_TO} \semicolon \terminal{NOT\_EQUAL\_TO} \semicolon \\
			\> \terminal{LIKE} \semicolon \terminal{IS\_NULL} \semicolon \\
			\> \terminal{IS\_NOT\_NULL} \semicolon \terminal{LESS\_THAN} \semicolon\\
			 \> \terminal{LESS\_THAN\_EQUAL\_TO} \semicolon \\
			 \> \terminal{GREATER\_THAN} \semicolon \\
			  \> \terminal{GREATER\_THAN\_EQUAL\_TO}\\ 
\nonterminal{Association} \arrow \= \nonterminal{RoleName} \semicolon \\
                                         \> \nonterminal{RoleName} \nonterminal{Association} \semicolon\\
                     			\> \nonterminal{RoleName} \nonterminal{Attribute} \semicolon  \\
						\> \nonterminal{RoleName} \nonterminal{Group} \\  
\nonterminal{Value} \arrow \nonterminal{xsd:string} \\
\nonterminal{QueryModifier} \arrow \= \nonterminal{DistinctAttribute} \semicolon\\
			\> \nonterminal{DistinctAttribute} \comma \nonterminal{AttributeNames}\\              
\end{tabbing}
\end{minipage}
\end{tabular}
\caption{CQL query context-free grammar \label{table:cqlgrammar}}
\end{table}

So, CQL traverses the UML class diagram graph, where the {$\langle$}Target{$\rangle$} is the initial class, the {$\langle$}Association{$\rangle$} conditions allow for path navigation by traversing sequences of consecutive classes and {$\langle$}Attribute{$\rangle$} conditions apply locally to individual classes. The terminal symbols {$\langle$}Group{$\rangle$} and {$\langle$}Group1{$\rangle$} represent the combination of two or more constraints over a particular node in the UML class graph.

%% file: ontoqueries.tex
As shown before, the caGrid queries rely on the structure of the underlying data resources, \ie their UML models. Thus, a biomedical researcher interested in retrieving data about, for example, a particular gene of interest will need to explore the UML model of each relevant data service and build a query considering the specific attributes and associations of the class maintaining the \emph{Gene} objects. The queries can be built programmatically or also through the caGrid portal\footnote{\href{http://cagrid-portal.nci.nih.gov}{http://cagrid-portal.nci.nih.gov}}, which allows to explore the UML models and provides a query builder based on these models.

In this work, we propose a system that allows the user to concentrate on the concepts from the domain, as represented by the NCIt ontology on cancer, and build the ontology-based queries which are \emph{high-level}\footnote{By a high-level query, we mean a query that can be written without specific details about the structure of the target resource.} and \emph{descriptive}\footnote{By a descriptive query, we refer to queries that provide the criteria for the desired data rather than the procedure to find the data.}. Thus, the ontology-based queries can be applicable to any of the underlying data resources.

Apart from the cancer concepts found on NCIt, the queries combine elements from an ontology we built with metadata on UML models\footnote{We will see later, than the queries could also use elements from the list ontology\cite{drummond:putting06}.}, namely the \emph{UML model} ontology. This ontology contains OWL classes to represent UML classes and attributes (\emph{UMLClass}, \emph{UMLAttribute}), OWL object properties to represent UML associations and the relationship between a UML class and its attributes (\emph{hasAssociation}, \emph{hasAttribute}) and a data property to represent the values of attributes (\emph{hasValue}).

Some simple example queries\footnote{The example queries are given in Manchester OWL Syntax and are just intended to show queries retrieving objects from a UML class, a UML class with a condition over an attribute and two associated UML classes with a restriction over an attribute of one of the classes, respectively.} are:
\begin{inparaenum}[\itshape a\upshape)]
\item \emph{Specimen} to retrieve all the objects that are annotated with the Specimen concept;
\item \emph{Gene and hasAttribute some (Gene\_Symbol and hasValue value "BRCA\%")} to find all the genes whose symbol starts with the string BRCA;
\item \emph{Single\_Nucleotide\_Polymorphism and hasAssociation some (Gene and hasAttribute some (Gene\_Symbol and hasValue value "TGFB1"))} to obtain all the \emph{SNPs} associated with the \emph{Gene} \emph{Transforming Growth Factor Beta 1}\cite{gonzalez-beltran:CBMS2009}.
\end{inparaenum} In our system, these queries could be submitted to any data service, and they will be converted to the specific CQL query.

We note that the third query specifies SNPs that are associated with genes. This association may be present in different ways in two separate UML models. For example, the two corresponding classes may have a direct UML association, or the association may arise by traversing an association path from the first class to the second one. In order for our system to deal with those paths of associations, without the user requiring to know the specific underlying UML model, we define the \emph{hasAssociation} property as transitive and use reasoning to determine the paths.

Next, we introduce our transformation from caGrid models to an OWL2 representation and the query rewriting/translation approach, which transforms ontology-based queries into CQL queries. The OWL2 ontologies provide an unified view of the UML models and their semantic annotations, which allows us to apply reasoning over them.

%% file: owlrep.tex
\begin{floatingfigure}[r]{58mm}
  \begin{center}
    \includegraphics[width=56mm]{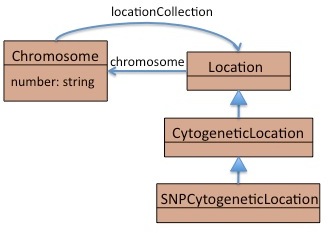}
  \end{center}
  \caption{Part of UML class diagram for caBIO 4.2 \label{figure:uml}}
\end{floatingfigure}

{\bf OWL model of UML class diagrams.} First, we present our customised UML-to-OWL transformation. This transformation differs from previous approaches, as explained in Section \ref{section:relatedwork}. Next, we describe the transformation and use the portion of the caBIO 4.2 information model in Figure \ref{figure:uml} to give examples. Every UML element is related to its counterpart in the \emph{UML model} ontology: all UML classes and attributes are defined as subclasses of \emph{UMLClass} and \emph{UMLAttribute}, respectively (see equations Eq. \ref{eq:umlclass} and Eq. \ref{eq:umlattribute} below\footnote{The prefixes used in the equations are: \emph{c:} for the caBIO 4.2 ontology, \emph{u:} for the UML model ontology, \emph{n:} for the NCIt ontology and \emph{l:} for the list ontology. We note that the name of an OWL class corresponding to an attribute includes the class name to avoid duplications and for associations, it includes its domain and range.}); all the UML associations are sub-properties of \emph{hasAssociation} (Eq. \ref{eq:hasAssociation}), and the datatype property \emph{hasValue} is used to specify the type of the attributes (Eq. \ref{eq:datatype}) as an existential restriction. Contrary to other UML-to-OWL transformations, we represent UML attributes as OWL classes. This is required so that the ontology-based queries can include the concepts associated with attributes.

\vspace{-6mm}
\begin{eqnarray}
\mbox{\small c:Chromosome} & \sqsubseteq  & \mbox{\small u:UMLClass} \label{eq:umlclass}\\[-1ex]
\mbox{\small c:Chromosome\_number} & \sqsubseteq & \mbox{\small u:UMLAttribute} \label{eq:umlattribute}\\[-1ex]
\mbox{\small c:Chromosome\_number} & \sqsubseteq & \exists \,\mbox{\small u:hasValue}.\mbox{\small xsd:string}\label{eq:datatype}\\[-1ex]
\mbox{\small c:Chromosome\_locationCollection\_Location} & \sqsubseteq & \mbox{\small u:hasAssociation} \label{eq:hasAssociation}
\end{eqnarray}

UML subclass and superclass relationships are represented with subsumption (Eq. \ref{eq:subsuperclass}). For each UML class, existential restrictions are added for its associations (Eq. \ref{eq:association}) and attributes (Eq. \ref{eq:attribute}). While UML does not explicitly represent inherited associations, our OWL representation makes them explicit, modeling the semantics of UML. For example, as the UML class \emph{Location} has an association \emph{chromosome} with the class \emph{Chromosome}, this association is inherited on the subclass \emph{CytogeneticLocation} (Eq. \ref{eq:inheritedassoc}).

\vspace{-6mm}
\begin{eqnarray}
\mbox{\small c:CytogeneticLocatoin} & \sqsubseteq &\mbox{\small c:Location} \label{eq:subsuperclass}\\[-1ex]
\mbox{\small c:Chromosome} & \sqsubseteq & \exists \, \mbox{\small c:Chromosome\_locationCollection\_Location}. \nonumber\\[-1ex]
&& \mbox{\small c:Location} \label{eq:association}\\[-1ex]
\mbox{\small c:Chromosome} & \sqsubseteq & \exists \, \mbox{\small u:hasAttribute}. \mbox{\small c:Chromosome\_number} \label{eq:attribute}\\[-1ex]
\mbox{\small c:CytogeneticLocation} & \sqsubseteq &\exists \, \mbox{\small c:Location\_chromosome\_Chromosome}.\nonumber \\[-1ex]
&& \mbox{\small c:Chromosome} \label{eq:inheritedassoc}
\end{eqnarray}

We note that the generated OWL ontologies belong to OWL2EL\cite{Cuenca-Grau:JWebSemantics2008}, an OWL2 profile specifically designed to allow for efficient reasoning with large terminologies, which is polynomial in the size of the ontology. While OWL2EL disallows universal quantification on properties, it does allow the inclusion of transitive properties. Thus, it is suitable for our UML-to-OWL transformation customised for the rewriting approach as outlined before. 

{\bf OWL Representation of the Semantic Annotations.}
Apart from representing the UML model, we also model its mapping to NCIt, as maintained in caDSR. Through the CDEs, UML elements are annotated with a primary concept, which indicates the meaning of the element. In turn, a list of qualifier concepts may be used to modify the primary concept, giving specific meaning. As OWL2 does not natively supports the representation of lists, we used Drummond \etalnospace's design pattern for sequences\cite{drummond:putting06}  to model primary concepts and qualifier lists. The following equations give some examples on how the semantic annotations of UML classes (Eq. \ref{eq:classannotation}) and attributes (Eq. \ref{eq:attributeannotation}) with a single concept are modelled. Equation \ref{eq:qualifierlist} models the class \emph{c:SNPCytogeneticLocation} as being a \emph{n:Location} qualified with \emph{l:Chromosome\_Band} and \\\emph{n:Single\_Nucleotide\_Polymorphism}.

\vspace{-6mm}
\begin{eqnarray}
\mbox{\small c:Chromosome} & \sqsubseteq &  \mbox{\small n:Chromosome} \label{eq:classannotation}\\[-1ex]
\mbox{\small c:Chromosome\_numer}  & \sqsubseteq & \mbox{\small n:Name} \label{eq:attributeannotation}\\[-1ex] 
\mbox{\small c:SNPCytogeneticLocation}  & \sqsubseteq & \mbox{\small n:Location} \sqcap  (\mbox{\small l:OWLList} \sqcap \nonumber  \\[-1ex]
&& \exists \mbox{\small l:hasContents}. \mbox{\small n:Chromosome\_Band} \sqcap  \\[-1ex]
&& \exists \, \mbox{\small l:hasNext}.(\mbox{\small l:OWLList} \sqcap  \nonumber\\[-1ex]
&&\exists \, \mbox{\small l:hasContents}.\mbox{\small n:Single\_Nucleotide\_Polymorphism})) \nonumber\label{eq:qualifierlist}
\end{eqnarray}

{\bf Module Extraction from NCI Thesaurus Ontology.}
The NCIt ontology is very large, as it provides a common vocabulary for the whole cancer domain\cite{hartel:JBI2005}. Each caGrid data service is, in general, concerned with data pertaining to more specific domains than the whole NCIt ontology. Thus, for each caGrid data service referring to a subset $\Sigma$ of the NCIt vocabulary, there is a subset of terms and relationships from NCIt that is \textit{relevant}, called a \textit{module} from the ontology\cite{sattler:module09}. The module $\mathcal{M}$ represents all knowledge about the terms of the \textit{signature} $\Sigma$. One of the approaches to\textit{relevance} is logic-based: the module $\mathcal{M}$ is relevant for the terms $\Sigma$ if all the consequences of the ontology that can be expressed over $\Sigma$ are also consequences of $\mathcal{M}$\cite{sattler:module09}. We follow that approach by Sattler \etal\cite{sattler:module09} and extract an NCIt module for each of the information models in caGrid. For succinctness and efficiency, this module is used, as opposed to the whole NCIt ontology, for the semantic annotations of UML models and subsequent reasoning. However, we observe that we removed the disjoint axioms from the NCIt modules, as we noted before\cite{gonzalez-beltran:CBMS2009,McCusker:BMCBioinformatics2009} using subsumption to represent UML class to concept mapping may result in inconsistent ontologies as the annotations for a single class may come from two high-level branches in NCIt that are declared as disjoint.

%% file: querytrans.tex
This section describes how an ontology-based query is rewritten and then translated, first to an intermediate optimisation language and then to the target CQL language. While the overall approach is similar to our previous work\cite{gonzalez-beltran:CBMS2009}, previously we relied completely on justifications\cite{kalyanpur:finding07} and now we have extensively improved the approach by dealing with each of the steps independently. We provide the output of each step for the third query from Section \ref{section:ontoqueries}.

{\bf Parsing.}
First, the user query is syntactically parsed. The query uses concepts from the NCIt, the UML model (see Section \ref{section:owlrep}) and the list ontologies\cite{drummond:putting06}.

{\bf UML Extraction.}
\input{umlextraction}

{\bf Data Values Extraction.}
As the generated ontologies do not contain instances, the semantic validation of the query, expressed as an OWL class expression, must ignore the data expressions. This step extracts the data expressions, which will be reinserted later on. This step results in {\footnotesize \textit{c:SNP and hasAssociation some (c:Gene and hasAttribute some (c:Gene\_symbol))}}.

{\bf Semantic Validation.}
We use a reasoner to check that the resulting query can be satisfied. If the query cannot be satisfied, subsequent rewriting of the query is halted.

{\bf Properties Path Finder.}
\input{pathfinder}

{\bf Data Values Addition.}
At this point, we can retrieve the data expressions removed earlier and re-insert them into the corresponding OWL classes, resulting in {\footnotesize \textit{c:SNP and hasAssociation some c:GeneRelativeLocation and hasAssociation some (c:Gene and hasAttribute some (c:Gene\_symbol and hasValue value "TGFB1"))}}.

{\bf OWL Expression to MCC Translation.}
\input{oce2mcc}

{\bf MMC to CQL Translation.}
\input{mcc2cql}

%% file: umlextraction.tex
The NCIt concepts in the query are translated into specific UML classes, by reasoning over the generated ontologies. Each concept is the super-class of a UML class or UML attribute, depending on their position on the query. Often, a single NCIt concept will correspond to many UML classes (or attributes) and, in such cases, each UML class is returned to form an individual query. Therefore, the outcome of the UML extraction is a combination of possible queries given the extracted UML classes or attributes. The outcome for our example query is: {\footnotesize \textit{c:SNP and hasAssociation some (c:Gene and hasAttribute some (c:Gene\_symbol and hasValue value "TGFB1")) }}.

%% file: pathfinder.tex
This step deals with the ontology corresponding to the UML model (the semantic annotations do not need to be considered any longer) and aims at finding the path of UML classes related through the transitive property \textit{hasAssociation}\footnote{We note that the ontology is compliant with the OWL2 EL profile, as OWL2 EL supports the use of transitive object properties. For more information, see \href{http://www.w3.org/TR/owl2-profiles/}{http://www.w3.org/TR/owl2-profiles/}}. The path finder rewrites the expression using non-transitive properties, corresponding to UML associations, by using an explanation generator\cite{kalyanpur:finding07} that retrieves the justification for two classes to be connected via the transitive property, and thus allowing to find the intermediate classes. The path finder may find more than one path between a set of nodes and, in such cases, will return each path as a combination of possible queries for user selection. One path for our example query is: {\footnotesize \textit{c:SNP and hasAssociation some c:GeneRelativeLocation and hasAssociation some (c:Gene and hasAttribute some (c:Gene\_symbol))}}.

%% file: oce2mcc.tex
No calculus or algebra has been defined for the object-oriented query language CQL. To provide a translation with CQL as target language, we use the monoid comprehension calculus (MCC), as it is a formal framework to support object queries optimizations\cite{fegaras:optimizing00}. Object queries involve collections (e.g. sets, lists, bags), whose semantics can be captured by monoid comprehensions (MC). In this paper, we only overview MCs and its use in our system\footnote{For more details, we refer the reader to \cite{fegaras:optimizing00} and \cite{gonzalez-beltran:CBMS2009}}. Our approach is similar to the work by Peim \etal\cite{peim:query02}, but while they use an expansion algorithm to rewrite an OWL expression based on a set of acyclic set of definitions, we follow the specific steps described above.

A MC takes the form $\oplus \{e \talloblong \overline{q}\}$, where $\oplus$ is a monoid\footnote{A monoid of type $T$ is an algebraic structure defined by $(\oplus, Z_{\oplus})$ where $\oplus: T \times T \rightarrow T$ is an associative funcion and $Z_{\oplus}$ is the left and right identity of $\oplus$. A collection monoid is a monoid for a collection type (e.g. lists or bags) and must also specify a unit function building a singleton collection.} operator called the \emph{accumulator}, $e$ is the \emph{header} and $\overline{q}=q_1,\ldots,q_n, n \geq 0$ is a sequence of \emph{qualifiers}. A qualifier can take the form of a \emph{generator}, $v \leftarrow e'$ with $v$ a range variable and $e'$ an expression constructing a collection, or a \emph{filter} predicate. The symbol $\cupplus$ denotes the accumulator for bags\footnote{For example, $\cupplus\{ x \talloblong x \leftarrow \{1 , 2\}\}$ is the monoid comprehension representing the bag $\{\{1, 2\}\}$. }. For an OWL class expression from the previous step, an MCC expression is built such that: the header variable is determined by the first concept in the query and the qualifiers are built for each of the remaining expressions. The MCC expression for our example is:
{ \footnotesize \textit{
$\cupplus$ \{ s $\talloblong$ s $\leftarrow$ SNP, r $\leftarrow$ s.relativeLocationCollection, r $\leftarrow$ GeneRelativeLocation, g $\leftarrow$ r.gene, g $\leftarrow$ Gene, g.symbol=TGFB1 \}
}}

%% file: mcc2cql.tex
Translating the MCC expression into CQL amounts to: define as \emph{Target} the type of the variable that appears in the header and then, including an \emph{Association} per each pair of generators, one determining the name (the class to which they belong) and the other identifying the role name; include an
\emph{Attribute} restriction for each filter. As this last step is the only one involving CQL, only this last step requires to be modified to extend our methodology to other model-driven architectures with a different target language. The resulting CQL in the example is:

\lstset{basicstyle=\scriptsize}
\begin{lstlisting}[label=cqlquery,showstringspaces=false]{float}
<ns1:CQLQuery xmlns:ns1="http://CQL.caBIG/1/gov.nih.nci.cagrid.CQLQuery">
<ns1:Target name="gov.nih.nci.cabio.domain.SNP">
  <ns1:Association name="gov.nih.nci.cabio.domain.GeneRelativeLocation" 
  roleName= "relativeLocationCollection">
   <ns1:Association name="gov.nih.nci.cabio.domain.Gene" roleName="gene">
    <ns1:Attribute name="symbol" predicate="EQUAL_TO" value="TGFB1"/>
   </ns1:Association> 
   </ns1:Association>
</ns1:Target>
 </ns1:CQLQuery>
\end{lstlisting}

\comment{

Additionally, CQL can contain Query Modifiers, indicating to return a 'count only' result, 
the whole objects, a single Distinct Attribute or a selection of Attribute Names for the target object. 
These modifiers cannot be expressed in the DL query. The translation from DL-query to CQL, via MCC, 
will consider retrieving full objects as default. Before submitting the resulting CQL query, 
the user will be asked to set the query modifiers.

CQL has some known caveats\footnote{\url{http://cagrid.org/display/dataservices12/CQL}}: it retrieves 
just objects of the Target type (not even subclasses of the Target type), only attributes with simple 
XML schema types are allowed to restrict the query or as return values, the association objects can be
used to impose restrictions on the result type but cannot be returned.

Translating the MCC expression into CQL amounts to:
include as Target in CQL the expression for the variable
in the head; include an Association per each pair of generators,
one determining the name (the class to which they
belong) and the other identifying the role name; include an
Attribute restriction for each filter. 
}

%% file: implementation.tex
We have implemented two modules, with the functionalities:
\begin{inparaenum}[\itshape a\upshape)]
\item an OWL generator, which transforms a caGrid annotated UML model into an OWL ontology and includes the generation of a module from the NCIt containing the concepts relevant to the UML model;
\item a query translation component, which takes as input a OWL class expression using concepts from the NCI thesaurus and transforms it into a CQL for a single data service.
\end{inparaenum}

For the first module, we also produced a caGrid analytical service called OWLGenService\footnote{The OWLGenService is accessible through the caGrid portal at \href{http://cagrid-portal.nci.nih.gov}{http://cagrid-portal.nci.nih.gov} and available at \href{http://stylus\_157.stylusinternet.net:9600/wsrf/services/cagrid/OwlgenService}{http://stylus\_157.stylusinternet.net:9600/wsrf/services/cagrid/OwlgenService}}, which provides a simple API for the extraction of modules from NCIt and for the ontology generation, given a specific information model.

The implementation was done in Java and uses caGrid version 1.3\footnote{\href{http://wiki.cagrid.org/display/caGrid13/Home}{http://wiki.cagrid.org/display/caGrid13/Home}}, the OWLAPI version 3.1.0\footnote{\href{http://owlapi.sourceforge.net/}{http://owlapi.sourceforge.net/}} (after upgrading from OWLAPI version 2), and relies on the reasoners Pellet 2.2.2\footnote{\href{http://clarkparsia.com/pellet/}{http://clarkparsia.com/pellet/}} and HermiT 1.3.0\footnote{\href{http://hermit-reasoner.com}{http://hermit-reasoner.com}}.

%% file: perfeval.tex
This section analyses the generated ontologies and presents two areas of performance evaluation that verify the viability of our approach.  
Since one important step in the query rewriting/translation process is the \textit{property path finder} (see Section \ref{section:querytrans}), we firstly introduce some metrics to assess the paths in the generated ontologies. These paths are sequences of concepts linked by object properties. Secondly, we present the generation times for the module extraction, the ontology generation and the inference of the ontologies using both the Pellet and HermiT reasoners. These results show that the generation of the ontologies that make possible our approach is done in a timely fashion. Thirdly, we evaluate the performance of the query rewriting process, showing a breakdown of the constituent parts of the rewriting algorithm. For this evaluation, we considered two sets of five queries each run over the caBIO data service\footnote{\href{http://cabiogrid42.nci.nih.gov:80/wsrf/services/cagrid/CaBIO42GridSvc}{http://cabiogrid42.nci.nih.gov:80/wsrf/services/cagrid/CaBIO42GridSvc}}, where each set consists of queries that involve paths of lengths one and two. The tests were run on a Red Hat Enterprise Linux Server release 5.3 (Tikanga) with 64 bits and 48285 MB of RAM.

This section analyses the generated ontologies and presents two areas of performance evaluation that verify the viability of our approach.  
Since one important step in the query rewriting/translation process --- from Section \ref{section:querytrans} --- is the \textit{property path finder}, we firstly introduce some metrics to assess the sequences of concepts linked by object properties (paths) in the generated ontologies. Secondly, we present the generation times for the module extraction, the ontology generation and the inference of the ontologies using both the Pellet and HermiT reasoners. These results show that the generation of the ontologies that make possible our approach take a short time. Thirdly, we evaluate the performance of the query rewriting process with a breakdown of the constituent parts of the algorithm. For this evaluation, we considered two sets of five queries each, where each set consists of queries that involve paths of lengths one and two. The results were obtained by running on a Red Hat Enterprise Linux Server release 5.3 (Tikanga) with 64 bits and 48285 MB of RAM. 

Throughout this section, we have grouped caGrid projects into three distinct subsets: projects that are available from the \emph{caDSR} service, all data services that are registered with the \emph{caGrid} default index service\footnote{\href{http://cagrid-index.nci.nih.gov:8080/wsrf/services/DefaultIndexService}{http://cagrid-index.nci.nih.gov:8080/wsrf/services/DefaultIndexService}}, and \emph{Information Models} (or \emph{InfoModels}) (those models that are supported by a deployed service from the \emph{caGrid} Index Service)\footnote{It should be noted that not all caDSR projects are included in the metrics; some contained errors (their semantic metadata is not complete or refers to an older version of the NCI thesaurus) and some models are targeted for data modelling, rather than specifically holding data, making them not representative for our system. Out of the 136 projects in caDSR, 16 were excluded from the analysis for these reasons. However, none of the excluded projects had an associated service. Additionally, the \emph{caGrid} subset has 63 services and \emph{InfoModels} has 23 projects.}. We note that the groups \emph{caGrid} and \emph{InfoModels} are the more relevant for our system, as only against these projects it is possible to execute CQL queries. While \emph{InfoModels} include a single project from caDSR for a set of deployed services corresponding to that project, \emph{caGrid} may include the results for several services that correspond to a single model. Thus, the \emph{caGrid} results will be skewed according to the relative weight of services as opposed to models. 


{\bf Analysis of the OWL representation of the information models.}
\input{owlmetrics}

\begin{figure}[h]
\begin{center}$
\begin{array}{cc}
\includegraphics[width=2.0in]{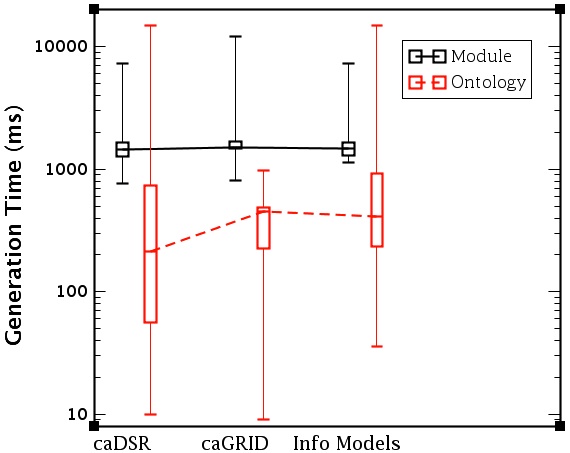} 
\includegraphics[width=2.0in]{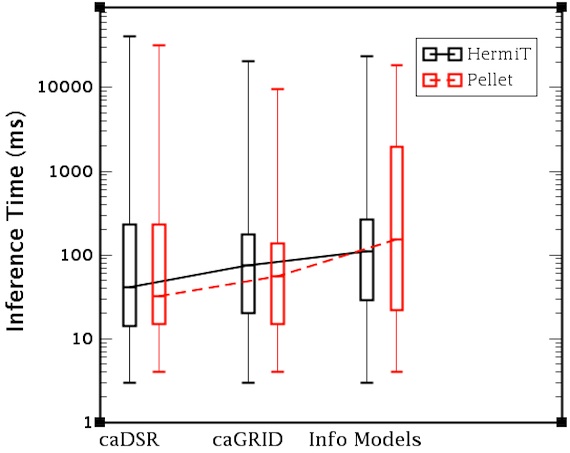}
\end{array}$
\end{center}
\caption{The generation and inference times. \label{fig:perfEval}}
\end{figure}

{\bf Ontology Generation, Module Extraction and Classification.}
In order to isolate any overhead caused by variations in network performance, we extracted the XML corresponding to each project (or information model) in caDSR. This is a preliminary step to run the performance evaluator locally, and we do not include any data about the performance of this stage. We generate four ontologies for each project: the NCIt module ontology (incorporating the concepts from NCIt relevant to the project), the annotated UML ontology (including the classes describing the UML model) and inferred versions of the two ontologies\footnote{We generate the inferred ontologies classifying the generated ontologies using both the HermiT and Pellet reasoners.}. We recorded the time for each generation and Figure \ref{fig:perfEval} shows them for the four ontologies per project grouped by subset. The times are presented in a logarithmic scale. We can see that the vast majority (75\%) of NCIt modules take less than 2 seconds to generate and even less time for ontology generation. The classification of the generated ontologies is also timely, with the average inference of the Pellet and HermiT reasoners never taking longer than 100 milliseconds.

\begin{figure}[h]
\begin{center}$
\begin{array}{cc}
\includegraphics[width=1.6in]{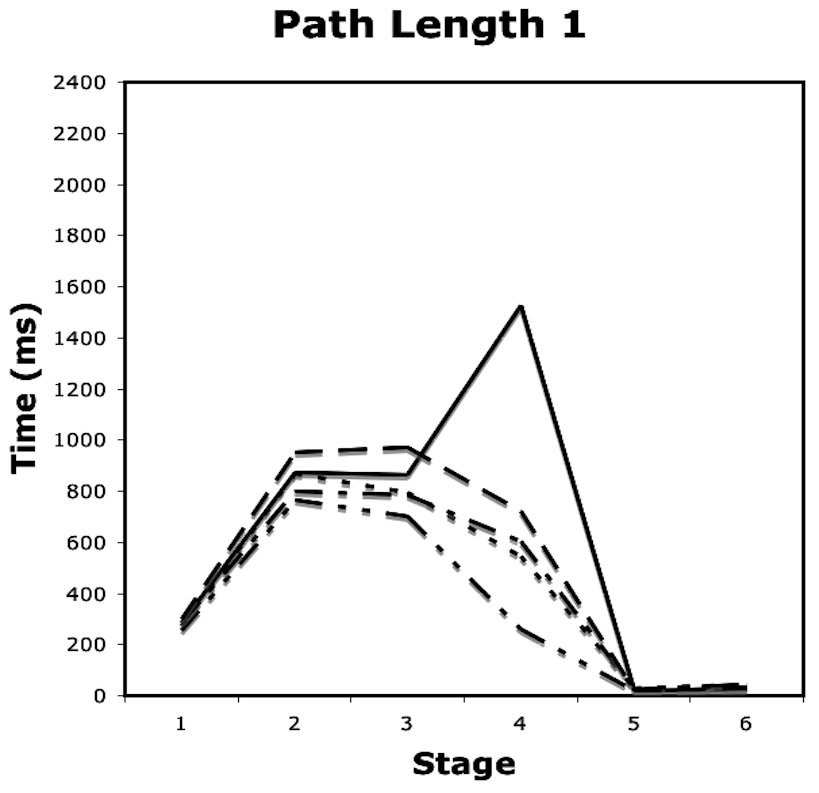} 
\includegraphics[width=1.6in]{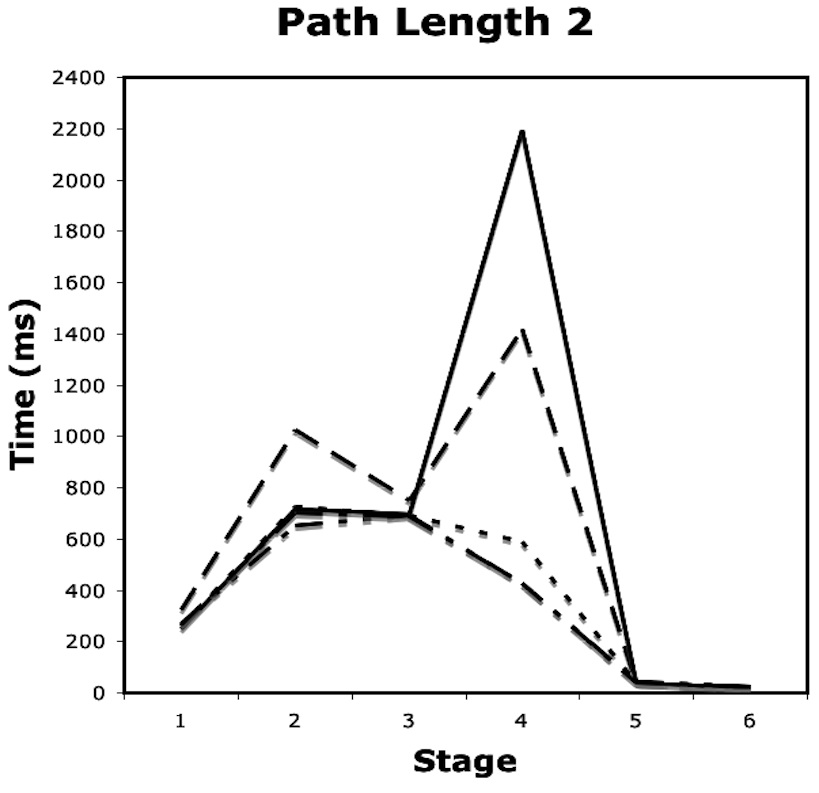}
\includegraphics[width=1.6in]{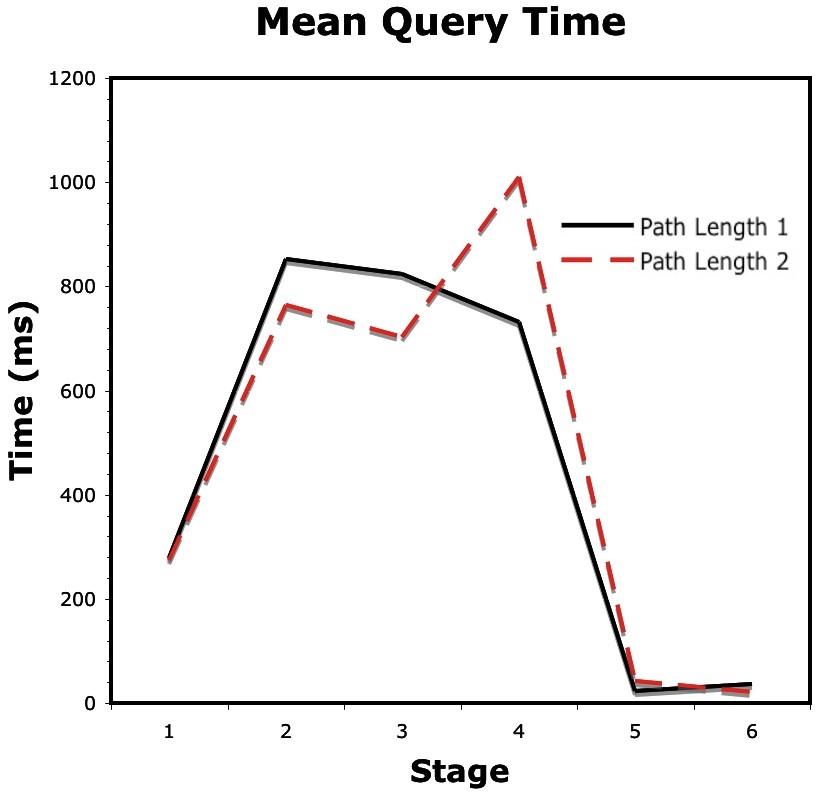}
\end{array}$
\end{center}
\caption{Query rewriting results at varying path lengths. \label{fig:qrMetric}}
\end{figure}

{\bf Query Rewriting Evaluation.}
We have developed a suite of queries of varying complexity in order to evaluate the query rewriting. The results are presented in figure \ref{fig:qrMetric}, which shows the average time\footnote{Each query was ran 5 times and the average times calculated.} taken at each stage of the query rewriting process\footnote{These correspond to the stages of query rewriting; parsing, UML extraction, validation, path finding, MCC conversion and CQL conversion. For more information, refer to section \ref{section:querytrans}.} for five queries whose rewriting has path length one, five queries whose rewriting has path length two and the mean times for these two sets. The query's path length refers to the number of intermediate nodes in the rewritten query. We can see from figure \ref{fig:qrMetric} that, while the path length has a marked effect on the time taken at the path finding stage, the other stages of implementation remain largely unaffected. We therefore maintain that, given our analysis of paths within our target ontologies described above, we can provide query rewriting in a timely and efficient manner.

%% file: owlmetrics.tex
While ontology metrics have been defined in several tools \cite{garcia:OWL10}, these have focused on basic metrics (e.g. number of classes) and semantic-based metrics (e.g. relationship richness) that allow for the comparison and quality evaluation of the ontologies. Here,  we will focus on the presentation of some bespoke metrics we developed to measure the proliferation and complexity of paths within the ontologies, as these will ensure the viability of our approach.

As seen in Section \ref{section:querytrans}, our rewriting process seeks to remove the upper-level and transitive object property \emph{hasAssociation} and express the query using only non-transitive properties, which correspond to the UML associations in the models. In order to achieve this, we consider the paths between pairs of concepts from the query connected through the \emph{hasAssociation} property. The calculation of these paths is not trivial; there may be many intermediate nodes and there may be more than one path for a given pair of concepts. We define a \emph{journey} as a traversal from one concept to another. A \emph{journey} may have one or many paths, which represent the possible routes that the traversal can take. Thus, it is important to evaluate these aspects of the ontologies in order to assess the viability of a rewriting tool.

We propose the following metrics as a measure of complexity in this respect. The \emph{Longest Path} is the maximum path length that may be computed within a given ontology. The longest path length provides an indication of the worse case for path calculation times. The \emph{Average Paths per Journey}  reflects the degree of path expansion within the rewriting algorithm, as each journey (e.g. from Node A to Node B) may have many different paths. The rewriting algorithm should return all possible paths as each path may refer to a different expression of the query. When we consider that a single query may include multiple independent journeys, the possible query rewritings can become very large. The \emph{Average Nodes per Path} is the average number of nodes that must be visited in order to return a single path. These metrics can affect the path calculation time as well as the complexity of the resulting query.

\begin{figure}[h]
\begin{center}$
\begin{array}{ccc}
\includegraphics[width=1.5in]{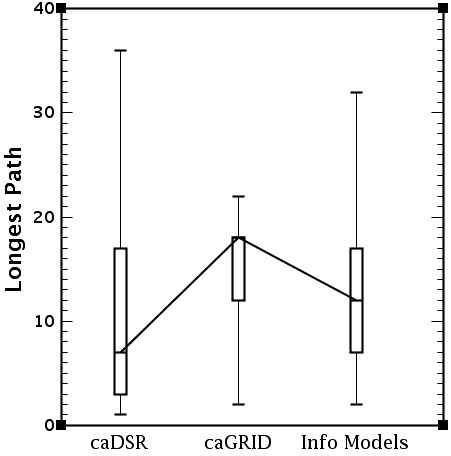} &
\includegraphics[width=1.5in]{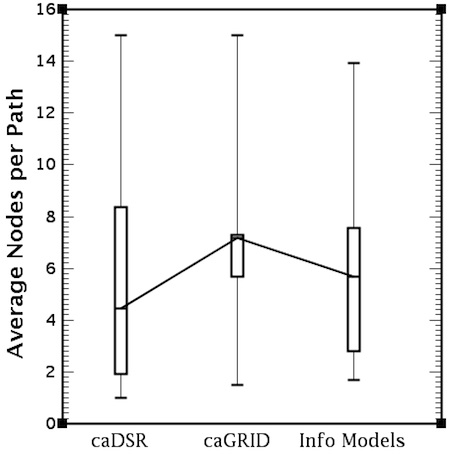}&
\includegraphics[width=1.5in]{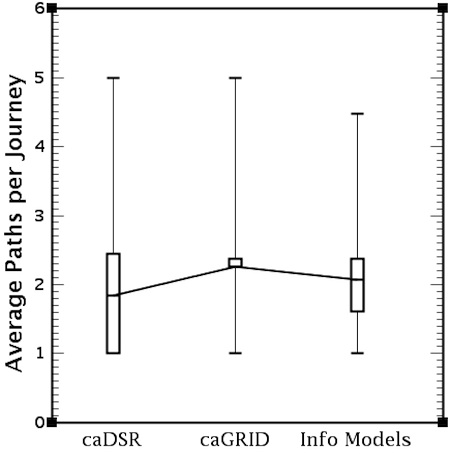}
\end{array}$
\end{center}
\caption{The Path Metrics. \label{fig:pathMetrics}}
\end{figure}

Figure \ref{fig:pathMetrics} illustrates three box-and-whisker plots with the results of the path metrics for each project subset. We observe that while the longest path can have up to 36 nodes, for 75 \% of the projects in each category their length is less than 17 or 18. The median of the average path length varies between 4 and 7 nodes over the three subsets, and for 75 \% of the \emph{InfoModels} the average path length is less than 8. The median of the average paths is around 2 paths per journey, and for 75 \% of the projects in each category the average path per journey is less than 2.5. This indicates that we will be returning a low number of path combinations as a result.  These results, then, verify that the paths within the ontologies are manageable and appropriate for our rewriting tool. We also note that in all the metric diagrams, the caGrid subset is often very densely clustered around the mean. This is due to the fact that there are often many caGrid services for the same project that differ to one another very slightly or not at all, which can result in multiple similar or identical results.

%% file: relatedwork.tex
The UML-to-OWL transformation has been studied in different contexts and applications varying from the detection of inconsistencies in UML diagrams to use as interchangeable modeling artifacts\cite{berardi:reasoning05,gasevic:mda07}. We have also provided different alternative transformations before\cite{gonzalez-beltran:CBMS2009,McCusker:BMCBioinformatics2009} and have improved it here so that the UML transformation conforms with OWL2EL profile, where the semantic annotations use subsumption and additionally, primary concepts and qualifiers are modelled as sequences.

The use of semantic web technologies to support semantic queries over distributed data environments in biomedicine have been implemented in systems such DartGrid\cite{chen:dartgrid06b} (for traditional chinese medicine), ACGT\cite{tsiknakis:semantic08} and semCDI\cite{shironoshita:semCDI08} (for cancer). To the best of our knowledge, the latest is the only work over the caGrid infrastructure. All these systems support SPARQL queries over the resources, while our system allows for high-level descriptive queries which do not need to be based on the structure of particular resources. Additionally, our approach using MCC as an intermediary language provides support for optimisations and generality, as a different target language could be used, even SPARQL.

%% file: conclusions.tex

This paper presented the design and implementation of an ontology-based querying functionality implemented over a service-oriented, model-driven infrastructure aimed at sharing cancer research data. In particular, the implementation was based on the caGrid infrastructure, but the approach could be used over similar model-driven software infrastructures. We presented:
\begin{inparaenum}[\itshape a\upshape)]
\item the generation of customised OWL2 ontologies from annotated UML models, based on the ISO11179 standard for metadata registries, which differs from traditional UML-to-OWL conversions and it is an improvement from \cite{gonzalez-beltran:CBMS2009}, mainly as we now generate OWL2EL ontologies for the UML models and support annotations with primary concept and qualifiers;
\item an analysis of the generated ontologies by determining several relevant and bespoke ontology metrics concerning paths and their complexity, which justify the viability of our query rewriting/translation technique;
\item a caGrid analytical service implementing the OWL Generation facility;
\item an analysis of the capabilities of the caGrid query language, and the queries it supports;
\item a significant revision and improvement of the query rewriting and translation steps to transform a domain ontology-based query into CQL;
\item an extensive performance evaluation of the OWL generation and module extraction, plus an assessment of the querying rewriting and translation process and its viability.
\end{inparaenum}
In future work, we will extend the query rewriting/translation evaluation providing a varied query set, explore the use of an OWL2EL reasoner to improve performance of the path finding process and support federated queries across data resources, where the selection of join conditions will be provided by a semantic analysis of the distributed resources.

%% file: acknowledgements.tex
\hspace{0.1cm}The authors are grateful to the National Cancer Research Institute Informatics Initiative for support for their research.